\documentclass[twoside,portrait,leqno,letterpaper,twocolumn,10pt]{article}
\usepackage{ltexpprt}

\usepackage{hyperref}
\usepackage{color}
\usepackage{graphicx}
\usepackage{amsmath,amssymb}
\usepackage{algorithm}
\usepackage{algorithmic}
\usepackage{multicol,multirow}
\usepackage{lscape}
\usepackage{geometry}
\usepackage{lscape}

\newif\ifBibtex\Bibtexfalse
\newif\ifDraft\Draftfalse

\makeatletter
\def\blfootnote{\gdef\@thefnmark{}\@footnotetext}
\makeatother

\ifDraft

\newcommand{\relocate}[1]{\textcolor{blue}{}}
\newcommand{\rewrite}[1]{\textcolor{red}{#1}}
\else

\newcommand{\relocate}[1]{}
\newcommand{\rewrite}[1]{#1}
\fi

\newcommand{\secref}[1]{Section \ref{#1}}
\newcommand{\figref}[1]{Fig.\ref{#1}}
\newcommand{\tabref}[1]{Table \ref{#1}}
\newcommand{\algref}[1]{Algorithm \ref{#1}}
\newcommand{\apxref}[1]{Supplementary Material Section \ref{#1}}

\begin{document}

\title{\Large Discriminative Prototype Set Learning for Nearest Neighbor Classification}
\author{Shin Ando\thanks{{School of Management},
{Tokyo University of Science} } \\
}
\date{\texttt{ando@rs.tus.ac.jp}}

\maketitle

\begin{abstract}
{The nearest neighbor rule is a classic yet essential classification model, particularly in problems where the supervising information is given by pairwise dissimilarities and the  
embedding function are not easily obtained.} 
Prototype selection provides means of generalization and improving efficiency of the nearest neighbor model, but many existing methods assume and rely on the analyses of the input vector space. 
In this paper, we explore a dissimilarity-based, parametrized model of the nearest neighbor rule. 
\rewrite{In the proposed model, the selection of the nearest prototypes is influenced by the parameters of the respective prototypes. It provides a formulation for minimizing the violation of the extended nearest neighbor rule over the training set in a tractable form to exploit numerical techniques.} 
We show that the minimization problem reduces to a large-margin principle learning 
and demonstrate its advantage by empirical comparisons with other prototype selection methods. 
\end{abstract}

{\noindent \bf keywords}: {Nearest neighbor rule, Prototype selection, Soft maximum, Large-margin principle}

\section{Introduction}
The nearest neighbor rule is one of the most widely used models for classification  \cite{Olvera-Lopez:2010:RIS:1833239.1833256,Wu:2009:TTA:1554380}
and essential in domains where the objects are characterized by pairwise  dissimilarities, e.g, protein structures and genome sequences in biochemistry and population or groups of people in economics and psychology.
The supervising information for such objects is often provided as a matrix of mutual  dissimilarities, which the nearest neighbor rule can directly exploit.

Many classification models require training samples characterized by a set of features and represented as vectors. It is thus necessary to find an embedding of objects onto a Euclidean space in order to exploit such models for dissimilarity data.
Such techniques have been studied in the context of multi-dimensional scaling \cite{Torgerson1965Psychometrika} and have been incorporated with discriminative learning principles such as structural risk minimization  \cite{Graepel:1998:CPP:3009055.3009117}.
Another example of this approach is the approximate nearest neighbor search (ANNS) \cite{Andoni:2008:NHA:1327452.1327494,Pauleve:2010:LSH:1814586.1814748}, which employs locality sensitive hashing (LSH) for a probabilistic embedding of objects onto a low-dimensional space. 
These techniques, however, are respectively subject to some assumptions regarding, e.g., the properties of the embedded space or the functions used to compute dissimilarities.

The nearest neighbor classification model, on the other hand, is quite unconstrained, but
the issues of memory and computational efficiency for finding nearest neighbors are critical in practice \cite{Brighton:2002:AIS:593433.593527}.
\emph{Prototype selection} methods provide the means to reduce the time and space complexity for nearest neighbor computation and to generalize the nearest neighbor rule, by selecting a set of prototypes from the full training data for the prediction on the test data.  
\rewrite{Although there has been a substantial amount of literature on prototype selection published in the topic of data mining,} 
many such techniques exploit the analysis of the feature space, e.g., removing prototypes far from the decision boundary of two classes to reduce redundancy or conversely, removing those near the boundary for generalization \cite{Triguero:2012:TES:2334811.2334873}. 
In addition, these models require the \emph{wrapper} technique and cross validation to empirically tune parameters  
 \cite{Olvera-Lopez:2010:RIS:1833239.1833256} which are computationally expensive, due to their intractable formulations.

In this paper, we revisit the prototype selection problem to explore a parametric approach based on a discriminative learning principle. 
\rewrite{The key intuition of the proposed approach is two-fold.} First, we define a parametric extension of the nearest neighbor rule, in which the choice of the nearest neighbor is adjusted by a numerical parameter assigned to each prototype. \rewrite{The parameter values are also used to infer whether the instances are relevant for prediction.} 
Secondly, \rewrite{we introduce a differentiable approximation of the condition that the nearest neighbor rule predicts the class of a training instance correctly,} and subsequently derive a constrained optimization problem for minimizing the violation of the condition with regards to the numerical parameters of the prototypes.
We further show that this problem \rewrite{reduces to a large-margin principle learning using a sparse representation of the relations between the neighboring instances.}

In summary, the contribution of this work is a prototype selection method, \rewrite{which (a) can exploit complex distance or dissimilarity functions and do not rely on the presence and the analysis of the vector space and (b) enables an optimization algorithm for learning the parameters to extend the nearest neighbor rule.}   
The rest of this paper is organized as follows. \secref{sec:Related Work} discusses the related work.
\secref{sec:PRPS} describes the proposed method
\secref{sec:Empirical Results} shows the results of our empirical study. \secref{sec:Conclusion} presents our conclusion.

\section{Related Work}\label{sec:Related Work}

\rewrite{The weaknesses of the nearest neighbor classifier in terms of large time and memory requirements respectively for computing and storing the similarities is well-known. 
One way to address this issue is to reduce the number of prototypes at the pre-processing phase. This problem has been referred to as prototype selection \cite{Garcia:2012:PSN:2122272.2122582}, data reduction \cite{Wilson:2000:RTI:343196.343200}, instance selection \cite{Garcia-Pedrajas:2009:CEC:1657491.1657497,Garcia-Pedrajas:2014:BIS:2657522.2658088,Marchiori:2010:CCN:1687045.1687136}, and template reduction \cite{Fayed:2009:NTR:1657543.1657555,Pkalska:2006:PSD:1220974.1221359}. 
The goal of the problem is to find a compact subset of the prototypes which maintains or possibly improve the generalization error of the nearest neighbor classification while decreasing the computational burden.}

The above methods typically employ operations such as \emph{editing}, which eliminates or relocates \emph{noisy} prototypes that cause misclassifications near the borders of different classes \cite{Brighton:2002:AIS:593433.593527,Wilson:2000:RTI:343196.343200} for the effects of smoothing and generalizing the decision boundaries.
\emph{Condensation} is another common operation, which discards prototypes far from the class borders in order to reduce the redundancy of the prototype set without affecting the decision boundary \cite{Angiulli:2007:FNN:1313048.1313207}. 

The above methods are generally implemented with the \emph{wrapper} or the \emph{filtering} techniques \cite{Olvera-Lopez:2010:RIS:1833239.1833256}. The wrapper technique evaluates and selects different sets of prototypes based on the performance of the nearest neighbor classifier. The filtering technique selects prototypes based on their individual scores, which are less expensive to compute than the performance measure.

\relocate{In recent years, Approximate Nearest Neighbor Search (ANNS) \cite{Andoni:2008:NHA:1327452.1327494} has drawn strong interest and various techniques have been developed for image classification.
ANNS uses a family of hash functions to obtain \emph{buckets} to which the prototypes are hashed, and when a test sample is \emph{queried}, its hash value is used to retrieve relevant prototypes from the bucket. 
The error bounds of the approximation are derived from the probabilities that points closer and further apart than designated distances are hashed to the same and different buckets, respectively.}
\relocate{In the recent survey \cite{Garcia:2012:PSN:2122272.2122582}, the prototype selection methods aimed at reducing the generalization error have achieved better accuracy than ANNS.} 
\relocate{Additionally, the implementation of ANNS requires a class of hash function that can satisfy the conditions for approximation, which is limited to few well-known distance functions.} 
\relocate{Finding a suitable LSH family for a domain-specific distance function is generally difficult and is not actively pursued.}
\relocate{This paper focuses on a prototype selection approach which requires only the mutual distance matrix as the input, thus is applicable for problems where ANNS approach and also many prototype selection approaches are not.}  
\section{Proposed Method}\label{sec:PRPS}

\subsection{Preliminaries}\label{subsec:preliminaries}
Let $\mathcal{X}=\{{x}_i\}_{i=1}^n$ denote the set of labeled instances and $\mathcal{Y}=\{y_i\}_{i=1}^n$ their class values. Instances are not necessarily represented as vectors, as the nearest neighbor prediction requires only similarities/dissimilarities from the prototypes to a new instance. 
{The nearest neighbor classifier, $h$, returns the class prediction on a test sample ${t}$ as follows.
\begin{equation}\small
h({t})=y_i:\mathop{\arg\min}\limits_{i\in\{1,\ldots,n\}}{d}({t},{x}_i)
\label{eq:h(x)=y_i}
\end{equation}
\noindent where ${d}(\cdot,\cdot)$ denotes the dissimilarity function.
A prototype selection algorithm selects a subset $\mathcal{Z}\subseteq\mathcal{X}$ with which the nearest neighbor classifier yields smaller generalization error and run-time.}

In order to evaluate a candidate prototype set in the training phase, one must avoid trivial cases where the target instance being classified is also one of the prototypes set, as the prototypes and the training instances originally come from the same set of instances, $\mathcal{X}$. 
In the following sections, we employ a setup similar to the \emph{leave-one-out} cross validation, such that the target instance is always removed from the prototype set.
That is, the prototypes for predicting the class of $x_i$ will be a subset of the {remainder} of the labeled instances, denote by $\mathcal{Z}_i=\mathcal{X}\setminus\{x_i\}$. 
\rewrite{For distinction, we refer to an instance as a prototype only when it is used as one of the reference objects for the nearest neighbor rule, in this paper. In turn, we refer to an instance as a training instance only when it is the subject of prediction.} 

\subsection{Soft-Maximum Function}
The \emph{soft maximum} \cite{Boyd:2004:CO:993483} is an approximation of the function $\max(\cdot,\cdot)$ as $\log\left(\exp(\cdot)+\exp(\cdot)\right)$. The approximation is convex and differentiable, which are desirable for numerical optimization. 

A natural extension of the soft maximum for $m$ variables $x_1, \ldots, x_m$ is given by
\begin{equation}\small
\mathcal{M}(x_1,\ldots,x_m)=\log\sum_{i=1}^m\exp({x_i})
\label{eq:M(x_1,...,x_m)=}
\end{equation}
\rewrite{The base of the exponential, $\beta$, is larger than 1 and 
adjusted for the scale of the input values.}

\subsection{Adjusted Nearest Neighbor Rule}\label{subsec:Rank-Adjusted Nearest Neighbor Rule}
\rewrite{This section introduces an extension of the nearest neighbor rule to parametrize the model and the prototype selection problem.}

\rewrite{Let $x$ denote a target instance and $\mathcal{Z}$ denote the set of prototypes.} We denote by \rewrite{$R(x,x_j;\mathcal{Z})$ the rank of $d(x,x_j)$ in the set $\left\{d(x,x_j)\right\}_{x_j\in\mathcal{Z}}$. The rank takes an integer value from $\{1,\ldots,\#\mathcal{Z}\}$ and a smaller integer is a higher rank.} 
The nearest neighbor rule for a target instance $x$ is given as follows.
\begin{itemize}
\item Compute distances $\{d(x,x_j)\}_{x_j\in\mathcal{Z}}$
\item Assign rank $R(x,x_j;\mathcal{Z})$ to all $x_j\in\mathcal{Z}$ based on their distances to $x$
\item Return the class of the highest-ranked $x_j$
\end{itemize}
\noindent We introduce a non-negative parameter $\alpha(x_j)$ for each prototype to compute the adjusted rank $R'$.
\begin{equation}\small
R'(x,x_j;\mathcal{Z})=R(x,x_j;\mathcal{Z})+\alpha(x_j)
\label{eq:R'(x,s;`S)=}
\end{equation}
The adjusted nearest neighbor rule generalizes the nearest neighbor rule and produces identical predictions when $\alpha(x_1)=\ldots=\alpha(x_n)$. 

Based on the adjusted ranks, the nearest prototype can be passed over for another prototype. We thus refer to $\alpha(x_j)$ as the degradation parameter of the prototype $x_j$. For brevity, we will use $\alpha_j$ as $\alpha(x_j)$ when it is clear from context.
One may correct individual misclassifications by assigning larger degradations to the related prototypes. That is, if $x_j$ and $x_k$ are respectively the highest and the second highest-ranked prototypes for a training case $(x,y)$ where $y_j\neq{y}$ and $y_k={y}$, \rewrite{the misclassification may be avoided by assigning a relatively larger degradation to $\alpha_j$ than $\alpha_k$.} 

\rewrite{The aim of this extension is to enable the training of the degradation parameters to improve the classification performance over all hold-out cases. Based on their values, we can identify which prototypes are more relevant for prediction and reduce the prototype set accordingly.} 
\rewrite{The motivation for focusing on ranks rather than distances is to avoid the issue of scaling. The latter may vary substantially in scale depending on the function, thus may require \emph{ad-hoc} adjustments.}

\subsection{Approximate Nearest Neighbor Rule}\label{subsec:Approximation of Nearest Neighbor Rule}

Let $(x_i,y_i)$ denote a pair of a hold-out instance and its class label. We denote by $\mathcal{P}$ the subset of prototypes $\mathcal{Z}_i$ with the same label as $y_i$ and $\mathcal{Q}$ that of labels other than $y_i$.
The condition that the nearest neighbor rule correctly predicts $y_i$ is that its nearest neighbor is an element of $\mathcal{P}$. That is, 
\begin{equation}\small
\footnotesize\max\left(\left\{-{d}(x_i,p)\right\}_{p\in\mathcal{P}}\right) >
\max\left(\left\{-{d}(x_i,q)\right\}_{q\in\mathcal{Q}}\right)
\label{eq:max<max}
\end{equation}
Note that the trivial prediction does not occur because $x_i$ is not an element of $\mathcal{P}$.

In an ideal prototype set, the condition \eqref{eq:max<max} is satisfied over all {training} cases for all ${x_i}\in\mathcal{X}$.
There may not exist such an ideal set in general, and thus a practical goal for selecting the prototypes is to reduce the cases where the conditions are violated, as much as possible. 

The formulation of such a principal is non-trivial due to the $\max$ \rewrite{function} in \eqref{eq:max<max}. Alternatively, we rewrite \eqref{eq:max<max} substituting the soft maximum $\mathcal{M}$ and the ranks of $d(x_i,q)$ and $d(x_i,p)$ for $\max$ and the distance values, respectively. 
\begin{eqnarray}
&&\mathcal{M}\left(\{-R(x_i,p;\mathcal{Z}_i)\}_{p\in\mathcal{P}}\right) 
\label{eq:M<M}\label{eq:M<M}\geq\\ &&
\mathcal{M}\left(\{-R(x_i,q;\mathcal{Z}_i)\}_{q\in\mathcal{Q}}\right)+1
,\forall{x_i}\in\mathcal{X}
\nonumber
\end{eqnarray}
In essence, \eqref{eq:M<M} describes the same condition as \eqref{eq:max<max} with regards to the ranks of the prototypes of the same and different classes in the hold-out case. \rewrite{The constant 1 is added to the RHS given that $R$ is the rank taking an integer value.} 

Substituting 
\eqref{eq:M(x_1,...,x_m)=} to \eqref{eq:M<M}, we have
\begin{eqnarray}
&&\log\left(\sum_{p\in\mathcal{P}}
\exp\left({-R\left(x_i,p;\mathcal{Z}_i\right)}\right)
\right)\\
&\geq&\log\left(\sum_{q\in\mathcal{Q}}
\exp\left(-R\left(x_i,q;\mathcal{Z}_i\right)\right)
\right)+1
\nonumber
\end{eqnarray}
Taking the exponential on both sides, we obtain
\begin{eqnarray}
&&\sum_{p\in\mathcal{P}}\exp\left(-R(x_i,p;\mathcal{Z}_i)\right)\\
&&-\sum_{q\in\mathcal{Q}}
\exp\left({-R\left(x_i,q;\mathcal{Z}_i\right)}\right)
\geq \rho(x_i)\nonumber
\label{eq:sum_p_in_P_exp(-R(x_i,p;Z_i))}
\end{eqnarray}
where $\rho(x_i)$ is 
\begin{equation}\small
\rho(x_i)=(\beta-1)\sum_{q\in\mathcal{Q}}
\exp\left({-R\left(x_i,q;\mathcal{Z}_i\right)}\right) 
\label{eq:rho(x_i)=}
\end{equation}
\noindent
Rewriting the LHS of \eqref{eq:sum_p_in_P_exp(-R(x_i,p;Z_i))}
given $\mathcal{P}\cup\mathcal{Q}=\mathcal{Z}_i$,
\begin{equation}\small
\sum_{x_j\in\mathcal{Z}_i}
\delta(y_i,y_j)\exp\left({-R\left(x_i,x_j;\mathcal{Z}_i\right)}\right)
\geq\rho(x_i)
\label{eq:sum_V_i`delta}
\end{equation} 
where 
$\displaystyle
\delta(y_i,y_j)=\left\{\begin{matrix}
1&\text{if\,}y_i=y_j\\
-1&\text{otherwise}
\end{matrix}\right.$\\[0.2em]

\noindent
Considering \eqref{eq:sum_V_i`delta} 
a constraint of nearest neighbor rule 
for each $x_i$, we introduce a slack variable $\xi_i$ to
represent the violation.
\begin{eqnarray}
 &&
\sum_{x_j\in\mathcal{Z}_i}
\delta(y_i,y_j)\exp\left({-R\left(x_i,x_j;\mathcal{Z}_i\right)}\right)
 \label{eq:sum_x_j_Z_delta}\\
 &&
\phantom{===}\geq\rho(x_i)-\xi_i
\nonumber
\end{eqnarray}
Each $\xi_i$ takes a non-negative value, and the violation of the nearest neighbor rule over the training set is given as $\sum_{x_i\in\mathcal{X}}\xi_i$.

\subsection{Problem Formulation}
Combining the rank degradation parameter and the soft-max approximation, 
we formulate a constrained optimization problem for the parametrized nearest neighbor classification model.

Rewriting \eqref{eq:sum_x_j_Z_delta} with the adjusted rank, 
\begin{eqnarray}
&&\sum_{x_j\in\mathcal{Z}_i}
\delta(y_i,y_j)\exp\left({-R\left(x_i,x_j;\mathcal{Z}_i\right)}
-\alpha(x_j)\right)\label{eq:x_j`V_i=delta(y_i,y_j)..}\\
&&\geq\rho'(x_i)-\xi_i
\nonumber
\end{eqnarray}
where $\rho'(x_i)$ is the residual term that corresponds to $\rho(x_i)$ in \eqref{eq:sum_x_j_Z_delta}, i.e., 
\begin{eqnarray}
\rho'(x_i)&=&(\beta-1)\sum_{q\in\mathcal{Q}}
\exp\left({-R'\left(x_i,q;\mathcal{Z}_i\right)}\right)
\nonumber\label{eq:rho'_i=}\\
&=&(\beta-1)\sum_{q\in\mathcal{Q}}
\exp\left(-R\left(x_i,q;\mathcal{Z}_i\right)
-\alpha(q)\right)
\nonumber
\end{eqnarray}
\eqref{eq:x_j`V_i=delta(y_i,y_j)..} is not a desirable form for a constrained optimization problem because the parameter being trained appears on both sides. That is, $\alpha(x_j)$ on the LHS and $\alpha(q)$ in $\rho'(x_i)$. 
Alternatively, we replace $\rho'(x_i)$ with $\rho(x_i)$, which is constant for each $x_i$, and the solution that satisfies \eqref{eq:sum_x_j_Z_delta} will also satisfy \eqref{eq:x_j`V_i=delta(y_i,y_j)..} since $\rho'(x_i)\leq{\rho(x_i)}$. For brevity, we denote $\rho(x_i)$ as $\rho_i$ when it is clear in the context.

Secondly, we consider the regularization for the degradation parameters.
\rewrite{By definition, the nearest neighbor rule with adjusted ranks behaves the same when all parameters are increased or decreased by the same amount,} which is problematic for convergence. The issue can be addressed by the regularization on $\{\alpha(x_j)\}$, to promote less adjustments of ranks given the same amount of violations. 

Let $\lambda$ denote the regularization function on $\{\alpha(x_j)\}$ and $C$ the trade-off coefficient. The constrained optimization problem is written as
\begin{equation}\small
\mathop{\arg\min}_{\alpha_j\geq0,\xi_i\geq0}\lambda(\alpha_1,\ldots,\alpha_n) + C\sum_{i=1}^n\xi_i 
\label{eq:p1objective}
\end{equation}
{subject to } 
\begin{eqnarray}
&&\sum_{x_j\in\mathcal{Z}_i}
\delta(y_i,y_j)\exp\left({-R\left(x_i,x_j;\mathcal{Z}_i\right)}-\alpha_j\right)
\label{eq:p1constraint}\\
&&\geq\rho_i-\xi_i,\,\forall{x_i}\in\mathcal{X}\nonumber
\nonumber
\end{eqnarray}

To gain further insight on the above problem, 
we rewrite the LHS of the constraints 
as a linear combination of $\mathbf{w}=(w_1,\ldots,w_n)$ and $\mathbf{r}_i=(r_{i1},\ldots{r}_{in})$, where $w_j=\exp(\alpha_j)$ and $r_{ij}=\delta(y_i,y_j)\exp(-R(x_i,x_j;\mathcal{Z}_i))$. 
Using the $\ell$-2 norm of $\mathbf{w}$,
whose elements monotonically increase with $\{\alpha_j\}$, 
as the regularizer, we obtain
\begin{equation}\small
\mathop{\arg\min}\limits_{{w}_i\geq0,\xi_i\geq0}
\|\mathbf{w}\|^2+C\sum_i{\xi_i}
\label{eq:argmin_w_i}\\
\end{equation}
subject to
\begin{eqnarray}
\mathbf{w}^\top\mathbf{r}_i\geq{\rho}_i-\xi_i
,&&\forall{x_i}\in\mathcal{X}
\label{eq:w^Tr}
\end{eqnarray}
In \eqref{eq:argmin_w_i}, 
the problem has reduced to a quadratic programming 
for learning the vector $\mathbf{w}$ of a max-margin linear classifier. In turn, $\mathbf{r}_i$ is interpreted as a mapping of the hold-out case to a feature vector space. Each feature in $\mathbf{r}_i$ corresponds to a prototype, and characterizes the training instance $x_i$. 
\rewrite{The use of exponential with of $\mathbf{r}_i$ and the regularization on $\mathbf{w}$ induces a sparse representation to highlight the relevant prototypes.}  

Given the form of \eqref{eq:w^Tr}, 
the problem 
is solved by a constrained gradient descent \cite{kelley1999iterative}. 
The relevance of each prototype is scored by the learned degradation value, i.e., the highest adjusted rank for predicting the hold-out cases, 
\begin{equation}\small
s(x_j)=\alpha_j+\min\left(\{R(x_i,x_j;\mathcal{Z}_i)\}_{x_i\in\mathcal{Z}_j}\right)
\label{eq:s(x_j)=min}
\end{equation}
The prototypes with low scores are removed from the prototype set as they have less effect on the predictions.
If the size of the prototype set is fixed, the prototypes are selected by their scores from the highest to the lowest.
When the size of the prototypes is not given, one can choose the size of the set by cross validation. 
In the rest of the paper, we refer to the proposed method as Prototype Selection based on Rank-Exponential (REPS) and $\mathbf{r}_i$ as the Exponential Rank representation of $x_i$.

\section{Algorithm}\label{sec:Algorithm}
The procedure of REPS is briefly summarized as follows.
\begin{itemize}
  \item Compute the rank matrix from the input distance matrix
  \item Compute the parameter values in the constraints 
  \item Solve the quadratic programming problem regarding the degradation parameters  
  \item Eliminate candidates with high degradation parameters
\end{itemize}
{The description of the algorithm with more detail is shown in \algref{alg:1}.} 

\relocate{In order to address large datasets, we implemented 
a modification of the Structural SVM \cite{Joachims:2009:CTS:1612990.1613006} learning algorithm, for an efficient approximation algorithm for training SVM. 
It solves a modified SVM problem using the cutting-plane method, where $\rho_i$ is used as the relatively scaled margins of respective prototypes.  \algref{alg:2} in \apxref{apx:tables} shows the pseudo-code of the algorithm.}

\rewrite{With regards to the computational requirements, the space complexity is dominated by that of storing the mutual distance matrix among labeled instances, which is $O(n^2)$. The time complexity is dominated by the computation of ranks, which is $O(n \log n)$ for each instance.}

\begin{algorithm}[tb]
\caption{\rewrite{REPS algorithm}}\label{alg:1}
\begin{algorithmic}
\STATE {\bf INPUT}: Training set $\mathcal{X}=\{x_i\}_{i=1}^{n}$, regularization parameter $C$, selection size \rewrite{$k$} 
\STATE {\bf OUTPUT}: Prototype set $\mathcal{Z}\subset\mathcal{X}$
\STATE {\bf METHOD:}
\STATE Compute the distance matrix $\left[d(x_i,x_j)\right]_{n\times{n}}$ and the ranks $\{R(x_i,x;\mathcal{Z}_i)\}_{x\in\mathcal{Z}_i}$
\STATE Compute $r_{ij}=\delta(y_i,y_j)\exp\left(-R(x_i,x_j;\mathcal{Z}_i)\right)$
 and $\rho(x_i)$ \eqref{eq:rho(x_i)=}
\STATE Solve \eqref{eq:argmin_w_i} for $\mathbf{w}$
\FOR{$j=1$ to $n$}
\STATE Compute $\alpha_j=\log w_j$ and
  $s(x_j)$ from \eqref{eq:s(x_j)=min}
\ENDFOR
\RETURN 
$\mathop{\arg\max}\limits_{\mathcal{Z}\subset\mathcal{X}:\#\mathcal{Z}=k}\min(\{s(x_j)\}_{x_j\in\mathcal{Z}})$
\end{algorithmic}
\end{algorithm} 

\section{Empirical Results}\label{sec:Empirical Results}

This section presents an empirical study to evaluate REPS using public benchmark datasets. 
\subsection{Datasets}
For the first part of the experiment, we employed a collection of 34 datasets with vector features from the UCI Machine Learning Repository \cite{Bache+Lichman:2013} and the KEEL datasets \cite{Alcala-Fdez:2008:KST:1459241.1459248}.
The summary of the datasets is shown in \tabref{tab:summary_of_the_feature_vector_data}. 
\begin{table}[tb]
\caption{Dataset Properties 
}
\label{tab:summary_of_the_feature_vector_data}
\smallskip
\centerline{
\begin{tabular}{crrr}
\hline
 \text{} & \text{$\#$Ex.} & \text{$\#$Atts.} & \text{$\#$Cl.} \\
 \hline
 \text{appendicitis} & 106 & 7 & 2 \\
 \text{australian} & 690 & 14 & 2 \\
 \text{balance} & 625 & 4 & 3 \\
 \text{bands} & 539 & 19 & 2 \\
 \text{breast} & 286 & 9 & 2 \\
 \text{bupa} & 345 & 6 & 2 \\
 \text{cleveland} & 297 & 13 & 5 \\
 \text{contraceptive} & 1473 & 9 & 3 \\
 \text{crx} & 690 & 15 & 2 \\
 \text{dermatology} & 366 & 33 & 6 \\
 \text{ecoli} & 336 & 7 & 8 \\
 \text{german} & 1000 & 20 & 2 \\
 \text{glass} & 214 & 9 & 7 \\
 \text{haberman} & 306 & 3 & 2 \\
 \text{hayes-roth} & 160 & 4 & 3 \\
 \text{heart} & 270 & 13 & 2 \\
 \text{housevotes} & 435 & 16 & 2 \\
 \text{ionosphere} & 351 & 33 & 2 \\
 \text{iris} & 150 & 4 & 3 \\
 \text{lymphography} & 148 & 18 & 4 \\
 \text{mammographic} & 961 & 5 & 2 \\
 \text{monk-2} & 432 & 6 & 2 \\
 \text{movement$\_$libras} & 360 & 90 & 15 \\
 \text{newthyroid} & 215 & 5 & 3 \\
 \text{pima} & 768 & 8 & 2 \\
 \text{saheart} & 462 & 9 & 2 \\
 \text{sonar} & 208 & 60 & 2 \\
 \text{spectfheart} & 267 & 44 & 2 \\
 \text{tae} & 151 & 5 & 3 \\
  \text{vehicle} & 846 & 18 & 4 \\
  \text{wdbc} & 569 & 30 & 2 \\
  \text{wine} & 178 & 13 & 3 \\
  \text{wisconsin} & 699 & 9 & 2 \\
  \text{yeast} & 1484 & 8 & 10 \\
\hline
\end{tabular}}
\end{table}
\rewrite{Although some of the benchmarks are not large datasets, a substantial reduction of the prototype set 
is not irrelevant in a practical aspect, as the execution time for the nearest neighbor prediction is a considerable obstacle for its applications.} 
\relocate{In the second part of the experiment, we use a collection of 27 datasets from the UCR Time Series Dataset Repository \cite{UCR2015}. The input data for time series classification is given as a dissimilarity matrix of Dynamic Time Warping (DTW) \cite{BERNDT94V}, which is known to be highly effective in time series classification \cite{Ding:2008:QMT:1454159.1454226}. We tested window sizes between five and fifty for DTW, and the results were generally consistent among them. The window size is five for all results shown here.}  

\subsection{Baseline Methods}\label{subsec:Baseline Methods}
We selected five recent prototype selection methods, which ranked highly among the 42 algorithm reported in a recent survey \cite{Garcia:2012:PSN:2122272.2122582}, as baselines for comparative analysis. 
Class conditional instance selection (CCIS) \cite{Marchiori:2010:CCN:1687045.1687136}, Fast condensed nearest neighbor (FCNN) \cite{Angiulli:2007:FNN:1313048.1313207}. {Decremental Reduction Optimization (DROP3)} \cite{Wilson:2000:RTI:343196.343200}, Random Mutation Hill Climbing (RMHC) \cite{Department94prototypeand}. All baseline algorithms were executed in the KEEL software \cite{Alcala-Fdez:2008:KST:1459241.1459248}.
\relocate{Brief details of the baseline methods and their parameter specifications are 
presented in \apxref{apx:baselines}}
We also report the performance of the nearest neighbor classifier using the full training set as prototypes (NoPS).
Each method is evaluated using values in the second column and the best result is reported. With regards to the parameters of REPS, the regularization coefficient $C$ was set to $0.001$. Its results were robust with regards to $C$.

\subsection{Evaluation}\label{subsec:Evaluation}
\rewrite{The prototype selection methods are evaluated in two aspects: the accuracy of the nearest neighbor classifier and the compactness of the prototype set, and we use the error rate (ERR) and the selection rate (SLR) for respective measurements. ERR is the ratio of misclassifications with respect to the total number of predictions. SLR is defined as the ratio of the prototype set size with respect to the training set.}

Due to the trade-off between the two measures, however, it does not suffice to compare algorithms by single-objectives. Here, we employed \rewrite{two} approaches to compare the performance measures with the baselines in multi-objective manners: Pareto-rank and fixed selection rate error. \relocate{, and log odds ratio.} 
\relocate{We provide the details of the respective measurements in \apxref{apx:definitions}.}

For all measures, a smaller value indicates a better performance. The performances are averaged over 5-fold cross validation. \relocate{for the vector datasets.} 
The default training/test split is used for the time series datasets.
{Following the evaluations, we used the Wilcoxon's signed rank test to verify whether the difference between the proposed algorithm and each baseline algorithm is significant. }

\begin{landscape}
\begin{table}[h]
\caption{Summary of Error and Selection Rates (Vector Data)}
\label{tab:Error and Selection Rates Comparison}
\centerline{\footnotesize
\begin{tabular}{rclllllllllllll}
\hline
\multirow{2}{*}{ID} & \multirow{2}{*}{Data Name} & \text{NoPS} & \multicolumn{2}{c}{CCIS} & \multicolumn{2}{c}{FCNN} & \multicolumn{2}{c}{SSMA} & \multicolumn{2}{c}{DROP3} & \multicolumn{2}{c}{RMHC} & \multicolumn{2}{c}{REPS} \\
 & & \text{ERR} & \text{ERR} & \text{SLR} & \text{ERR} & \text{SLR} & \text{ERR} & \text{SLR} & \text{ERR} & \text{SLR} & \text{ERR} & \text{SLR} & \text{ERR} & \text{SLR} \\
\hline
1& \text{appendicitis} & 0.81 & {\textbf{0.18}} & {\textbf{0.031}} & 0.24 & 0.32 & {\textbf{0.15}} & {\textbf{0.038}} & 0.26 & 0.13 & {\textbf{0.13}} & {\textbf{0.094}} & 0.20 & 0.050 \\
2& \text{australian} & 0.69 & 0.44 & 0.030 & 0.41 & 0.32 & 0.41 & 0.014 & 0.39 & 0.11 & 0.42 & 0.10 & {\textbf{0.36}} & {\textbf{0.013}} \\
3& \text{balance} & 0.78 & 0.22 & 0.053 & 0.30 & 0.33 & {\textbf{0.12}} & {\textbf{0.030}} & 0.17 & 0.12 & {\textbf{0.12}} & {\textbf{0.10}} & {\textbf{0.28}} & {\textbf{0.020}} \\
4& \text{bands} & 0.71 & 0.41 & 0.054 & 0.42 & 0.48 & 0.42 & 0.056 & 0.39 & 0.32 & 0.42 & 0.099 & {\textbf{0.37}} & {\textbf{0.017}} \\
5& \text{breast} & 0.63 & 0.38 & 0.052 & 0.37 & 0.49 & 0.35 & 0.024 & {\textbf{0.33}} & {\textbf{0.20}} & 0.37 & 0.099 & {\textbf{0.34}} & {\textbf{0.022}} \\
6& \text{bupa} & 0.61 & 0.34 & 0.098 & 0.41 & 0.56 & {\textbf{0.33}} & {\textbf{0.057}} & 0.35 & 0.30 & {\textbf{0.33}} & {\textbf{0.098}} & {\textbf{0.45}} & {\textbf{0.017}} \\
7& \text{cleveland} & 0.53 & 0.74 & 0.31 & 0.68 & 0.61 & {\textbf{0.55}} & {\textbf{0.019}} & 0.61 & 0.17 & 0.60 & 0.098 & {\textbf{0.48}} & {\textbf{0.045}} \\
8& \text{contraceptive} & 0.43 & 0.53 & 0.23 & 0.53 & 0.71 & {\textbf{0.50}} & {\textbf{0.027}} & 0.52 & 0.27 & 0.50 & 0.099 & {\textbf{0.61}} & {\textbf{0.0064}} \\
9& \text{crx} & 0.78 & 0.47 & 0.031 & 0.44 & 0.31 & 0.40 & 0.016 & 0.44 & 0.12 & 0.42 & 0.10 & {\textbf{0.27}} & {\textbf{0.016}} \\
10& \text{dermatology} & 0.95 & {\textbf{0.61}} & {\textbf{0.047}} & 0.53 & 0.12 & {\textbf{0.80}} & {\textbf{0.036}} & 0.80 & 0.085 & 0.75 & 0.098 & {\textbf{0.073}} & {\textbf{0.048}} \\
11& \text{ecoli} & 0.79 & 0.32 & 0.14 & 0.24 & 0.37 & {\textbf{0.20}} & {\textbf{0.059}} & 0.26 & 0.16 & {\textbf{0.19}} & {\textbf{0.097}} & {\textbf{0.31}} & {\textbf{0.045}} \\
12& \text{german} & 0.67 & 0.49 & 0.040 & 0.42 & 0.50 & 0.42 & 0.026 & 0.41 & 0.22 & 0.39 & 0.10 & {\textbf{0.36}} & {\textbf{0.0092}} \\
13& \text{glass} & 0.72 & 0.55 & 0.14 & {\textbf{0.30}} & {\textbf{0.48}} & {\textbf{0.36}} & {\textbf{0.077}} & 0.40 & 0.25 & {\textbf{0.36}} & {\textbf{0.099}} & {\textbf{0.50}} & {\textbf{0.053}} \\
14& \text{haberman} & 0.65 & 0.36 & 0.034 & 0.34 & 0.51 & {\textbf{0.28}} & {\textbf{0.020}} & 0.35 & 0.20 & 0.31 & 0.098 & 0.33 & 0.027 \\
15& \text{hayes-roth} & 0.69 & 0.43 & 0.14 & {\textbf{0.36}} & {\textbf{0.49}} & {\textbf{0.41}} & {\textbf{0.066}} & 0.56 & 0.24 & 0.47 & 0.094 & {\textbf{0.46}} & {\textbf{0.052}} \\
16& \text{heart} & 0.79 & 0.40 & 0.043 & 0.41 & 0.37 & 0.42 & 0.026 & 0.44 & 0.17 & 0.37 & 0.097 & {\textbf{0.26}} & {\textbf{0.023}} \\
17& \text{housevotes} & 0.87 & {\textbf{0.35}} & {\textbf{0.018}} & 0.11 & 0.15 & {\textbf{0.079}} & {\textbf{0.026}} & 0.17 & 0.061 & 0.084 & 0.097 & {\textbf{0.11}} & {\textbf{0.019}} \\
18& \text{ionosphere} & 0.91 & {\textbf{0.36}} & {\textbf{0.017}} & 0.19 & 0.20 & {\textbf{0.10}} & {\textbf{0.033}} & 0.25 & 0.088 & 0.11 & 0.10 & 0.24 & 0.048 \\
19& \text{iris} & 0.85 & {\textbf{0.31}} & {\textbf{0.032}} & 0.41 & 0.13 & {\textbf{0.040}} & {\textbf{0.047}} & 0.087 & 0.073 & {\textbf{0.033}} & {\textbf{0.10}} & 0.053 & 0.15 \\
20& \text{lymphography} & 0.64 & 0.28 & 0.066 & 0.25 & 0.42 & {\textbf{0.24}} & {\textbf{0.057}} & 0.34 & 0.18 & 0.24 & 0.093 & 0.25 & 0.081 \\
21& \text{mammographic} & 0.80 & 0.44 & 0.017 & 0.29 & 0.39 & {\textbf{0.34}} & {\textbf{0.011}} & 0.38 & 0.14 & {\textbf{0.27}} & {\textbf{0.099}} & {\textbf{0.28}} & {\textbf{0.014}} \\
22& \text{monk-2} & 0.75 & 0.17 & 0.049 & 0.088 & 0.077 & {\textbf{0.037}} & {\textbf{0.031}} & 0.17 & 0.20 & 0.11 & 0.098 & 0.24 & 0.19 \\
23& \text{movement$\_$libras} & 0.76 & 0.39 & 0.31 & {\textbf{0.25}} & {\textbf{0.40}} & {\textbf{0.36}} & {\textbf{0.15}} & {\textbf{0.29}} & {\textbf{0.33}} & {\textbf{0.40}} & {\textbf{0.097}} & {\textbf{0.58}} & {\textbf{0.055}} \\
24& \text{newthyroid} & 0.85 & {\textbf{0.33}} & {\textbf{0.026}} & 0.14 & 0.12 & {\textbf{0.19}} & {\textbf{0.030}} & 0.15 & 0.12 & 0.12 & 0.099 & {\textbf{0.11}} & {\textbf{0.031}} \\
25& \text{pima} & 0.97 & 0.38 & 0.041 & 0.35 & 0.46 & {\textbf{0.28}} & {\textbf{0.026}} & 0.36 & 0.18 & 0.29 & 0.099 & {\textbf{0.32}} & {\textbf{0.010}} \\
26& \text{saheart} & 0.65 & 0.39 & 0.044 & 0.41 & 0.51 & 0.40 & 0.030 & 0.42 & 0.22 & 0.36 & 0.099 & {\textbf{0.31}} & {\textbf{0.017}} \\
27& \text{sonar} & 0.82 & {\textbf{0.46}} & {\textbf{0.046}} & 0.35 & 0.30 & {\textbf{0.32}} & {\textbf{0.074}} & 0.29 & 0.26 & {\textbf{0.28}} & {\textbf{0.096}} & {\textbf{0.28}} & {\textbf{0.11}} \\
28& \text{spectfheart} & 0.72 & 1.0 & 1.0 & 0.31 & 0.43 & {\textbf{0.21}} & {\textbf{0.024}} & 0.30 & 0.17 & 0.25 & 0.098 & {\textbf{0.20}} & {\textbf{0.024}} \\
29& \text{tae} & 0.58 & 0.59 & 0.16 & 0.58 & 0.60 & 0.64 & 0.083 & {\textbf{0.54}} & {\textbf{0.20}} & 0.57 & 0.099 & {\textbf{0.55}} & {\textbf{0.060}} \\
30& \text{vehicle} & 0.71 & 0.49 & 0.17 & {\textbf{0.38}} & {\textbf{0.49}} & {\textbf{0.45}} & {\textbf{0.065}} & {\textbf{0.43}} & {\textbf{0.24}} & 0.45 & 0.099 & {\textbf{0.63}} & {\textbf{0.0092}} \\
31& \text{wdbc} & 0.94 & {\textbf{0.21}} & {\textbf{0.0075}} & 0.077 & 0.12 & {\textbf{0.10}} & {\textbf{0.015}} & 0.11 & 0.044 & 0.077 & 0.099 & {\textbf{0.049}} & {\textbf{0.082}} \\
32& \text{wine} & 0.95 & {\textbf{0.29}} & {\textbf{0.025}} & 0.34 & 0.13 & 0.29 & 0.034 & 0.33 & 0.098 & 0.34 & 0.098 & {\textbf{0.073}} & {\textbf{0.12}} \\
33& \text{wisconsin} & 0.96 & 1.0 & 1.0 & 0.037 & 0.090 & {\textbf{0.031}} & {\textbf{0.0073}} & 0.062 & 0.025 & {\textbf{0.029}} & {\textbf{0.099}} & 0.045 & 0.012 \\
34& \text{yeast} & 0.53 & 0.55 & 0.22 & 0.48 & 0.66 & {\textbf{0.42}} & {\textbf{0.038}} & 0.48 & 0.23 & 0.42 & 0.099 & {\textbf{0.53}} & {\textbf{0.020}} \\
\hline
\end{tabular}}
 \end{table}
\end{landscape}
 
\subsection{Results}
\subsubsection{Vector Data}\label{subsubsec:Vector Data}
\begin{figure*}[tb]
\centerline{
\includegraphics[width=0.99\textwidth]{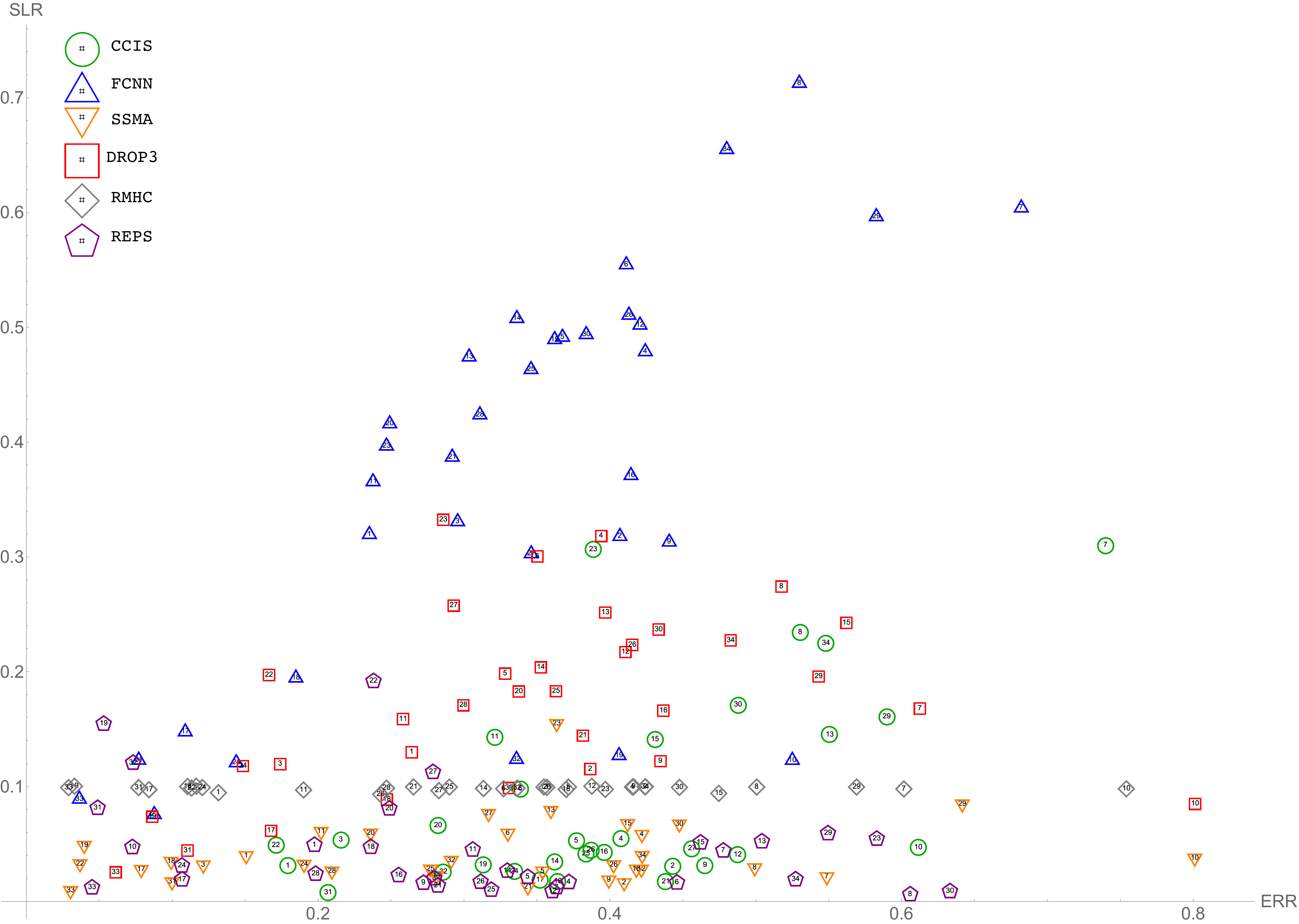}
}
\caption{Error and Selection Rates (Vector Data)}
\label{fig:Summary of Error and Selection Rates}
\end{figure*}

{\figref{fig:Summary of Error and Selection Rates} in illustrates the error and the selection rates of the evaluated algorithms. The $x$- and $y$-axes indicate the values of respective measures. Each marker represents the performances of one algorithm on one dataset. Each algorithm is distinguished by a unique color and shape of the marker. \rewrite{The numbers inside the markers identifies the dataset, which are same as the numbers shown in the first column of \tabref{tab:Error and Selection Rates Comparison}}.} 
\rewrite{Within the same problem (marker number), the markers generally lie from top-left to bottom right, due to the trade-off between the two measures. The shift toward the $x$-axis indicate the advantage in terms of compactness and the shift towards the $y$-axis indicate the advantage in accuracy.} 
\rewrite{Note that there is no intrinsic trade-off between the error and selection rates within the same method (marker shape), as the evaluation measures depend on the difficulty of the problems.} The balance of the two measures vary depending on the method, e.g., those of REPS and CCIS tend to lean toward higher selection rates.

The summary of the performances is presented in \tabref{tab:Error and Selection Rates Comparison}. Each row shows ERR and SLR of each method on one dataset. The selection rate for NoPS, which is always 1, is omitted. {The values of algorithms whose Pareto-rank is 1 are indicated in bold.} 
Overall, REPS and SSMA are Pareto-rank 1 in the largest numbers of datasets.

The FSR errors of the proposed algorithm and the error rates of the baselines are presented in \tabref{tab:Comparison of FSR Error Rates} 
The even number columns show the FSR errors of REPS using the same selection rate as the baseline in the next column. The odd numbered columns shows the ERR of the baselines. The selection rates of the baseline algorithms are the same as those shown in through 5$^\text{th}$-13$^\text{th}$ columns of \tabref{tab:Error and Selection Rates Comparison}. 

\begin{table*}[tb]
\caption{Summary of FSR Error Rates}
\label{tab:Comparison of FSR Error Rates}
\centerline{\footnotesize
\begin{tabular}{lllllllllll}
\hline
 \text{} & \text{REPS} & \text{CCIS} & \text{REPS} & \text{FCNN} & \text{REPS} & \text{SSMA} & \text{REPS} & \text{DROP3} & \text{REPS} & \text{RMHC} \\
\hline
 \text{appendicitis} & {$\mathbf{0.17}$} & 0.18 & {$\mathbf{0.17}$} & 0.24 & 0.22 & {$\mathbf{0.15}$} & {$\mathbf{0.19}$} & 0.26 & 0.19 & {$\mathbf{0.13}$} \\
 \text{australian} & {$\mathbf{0.28}$} & 0.44 & {$\mathbf{0.28}$} & 0.41 & {$\mathbf{0.34}$} & 0.41 & {$\mathbf{0.27}$} & 0.39 & {$\mathbf{0.27}$} & 0.42 \\
 \text{balance} & {$\mathbf{0.12}$} & 0.22 & {$\mathbf{0.15}$} & 0.30 & 0.21 & {$\mathbf{0.12}$} & {$\mathbf{0.14}$} & 0.17 & 0.15 & {$\mathbf{0.12}$} \\
 \text{bands} & {$\mathbf{0.32}$} & 0.41 & {$\mathbf{0.32}$} & 0.42 & {$\mathbf{0.38}$} & 0.42 & {$\mathbf{0.30}$} & 0.39 & {$\mathbf{0.36}$} & 0.42 \\
 \text{breast} & {$\mathbf{0.33}$} & 0.38 & {$\mathbf{0.32}$} & 0.37 & 0.43 & {$\mathbf{0.35}$} & {$\mathbf{0.32}$} & 0.33 & {$\mathbf{0.36}$} & 0.37 \\
 \text{bupa} & 0.39 & {$\mathbf{0.34}$} & {$\mathbf{0.39}$} & 0.41 & 0.40 & {$\mathbf{0.33}$} & 0.41 & {$\mathbf{0.35}$} & 0.40 & {$\mathbf{0.33}$} \\
 \text{cleveland} & {$\mathbf{0.48}$} & 0.74 & {$\mathbf{0.47}$} & 0.68 & {$\mathbf{0.49}$} & 0.55 & {$\mathbf{0.47}$} & 0.61 & {$\mathbf{0.46}$} & 0.60 \\
 \text{contraceptive} & 0.54 & {$\mathbf{0.53}$} & 0.55 & {$\mathbf{0.53}$} & 0.55 & {$\mathbf{0.50}$} & {$\mathbf{0.51}$} & 0.52 & 0.52 & {$\mathbf{0.50}$} \\
 \text{crx} & {$\mathbf{0.23}$} & 0.47 & {$\mathbf{0.23}$} & 0.44 & {$\mathbf{0.27}$} & 0.40 & {$\mathbf{0.26}$} & 0.44 & {$\mathbf{0.26}$} & 0.42 \\
 \text{dermatology} & {$\mathbf{0.076}$} & 0.61 & {$\mathbf{0.070}$} & 0.53 & {$\mathbf{0.076}$} & 0.80 & {$\mathbf{0.076}$} & 0.80 & {$\mathbf{0.076}$} & 0.75 \\
 \text{ecoli} & {$\mathbf{0.26}$} & 0.32 & 0.27 & {$\mathbf{0.24}$} & 0.27 & {$\mathbf{0.20}$} & 0.26 & {$\mathbf{0.26}$} & 0.27 & {$\mathbf{0.19}$} \\
 \text{german} & {$\mathbf{0.32}$} & 0.49 & {$\mathbf{0.32}$} & 0.42 & {$\mathbf{0.34}$} & 0.42 & {$\mathbf{0.31}$} & 0.41 & {$\mathbf{0.30}$} & 0.39 \\
 \text{glass} & {$\mathbf{0.36}$} & 0.55 & 0.34 & {$\mathbf{0.30}$} & 0.48 & {$\mathbf{0.36}$} & 0.42 & {$\mathbf{0.40}$} & 0.48 & {$\mathbf{0.36}$} \\
 \text{haberman} & 0.36 & {$\mathbf{0.36}$} & 0.35 & {$\mathbf{0.34}$} & 0.34 & {$\mathbf{0.28}$} & {$\mathbf{0.32}$} & 0.35 & 0.32 & {$\mathbf{0.31}$} \\
 \text{hayes-roth} & {$\mathbf{0.39}$} & 0.43 & 0.39 & {$\mathbf{0.36}$} & 0.44 & {$\mathbf{0.41}$} & {$\mathbf{0.38}$} & 0.56 & {$\mathbf{0.45}$} & 0.47 \\
 \text{heart} & {$\mathbf{0.20}$} & 0.40 & {$\mathbf{0.19}$} & 0.41 & {$\mathbf{0.21}$} & 0.42 & {$\mathbf{0.20}$} & 0.44 & {$\mathbf{0.21}$} & 0.37 \\
 \text{housevotes} & {$\mathbf{0.083}$} & 0.35 & {$\mathbf{0.086}$} & 0.11 & 0.11 & {$\mathbf{0.079}$} & {$\mathbf{0.094}$} & 0.17 & 0.087 & {$\mathbf{0.084}$} \\
 \text{ionosphere} & {$\mathbf{0.12}$} & 0.36 & {$\mathbf{0.13}$} & 0.19 & 0.26 & {$\mathbf{0.10}$} & {$\mathbf{0.21}$} & 0.25 & 0.20 & {$\mathbf{0.11}$} \\
 \text{iris} & {$\mathbf{0.073}$} & 0.31 & {$\mathbf{0.1}$} & 0.41 & 0.13 & {$\mathbf{0.040}$} & 0.1 & {$\mathbf{0.087}$} & 0.087 & {$\mathbf{0.033}$} \\
 \text{lymphography} & {$\mathbf{0.21}$} & 0.28 & {$\mathbf{0.22}$} & 0.25 & 0.24 & {$\mathbf{0.24}$} & {$\mathbf{0.23}$} & 0.34 & 0.26 & {$\mathbf{0.24}$} \\
 \text{mammographic} & {$\mathbf{0.21}$} & 0.44 & {$\mathbf{0.21}$} & 0.29 & {$\mathbf{0.30}$} & 0.34 & {$\mathbf{0.21}$} & 0.38 & {$\mathbf{0.23}$} & 0.27 \\
 \text{monk-2} & 0.18 & {$\mathbf{0.17}$} & 0.16 & {$\mathbf{0.088}$} & 0.25 & {$\mathbf{0.037}$} & 0.17 & {$\mathbf{0.17}$} & 0.19 & {$\mathbf{0.11}$} \\
 \text{movement$\_$libras} & {$\mathbf{0.32}$} & 0.39 & 0.32 & {$\mathbf{0.25}$} & {$\mathbf{0.35}$} & 0.36 & 0.34 & {$\mathbf{0.29}$} & {$\mathbf{0.39}$} & 0.40 \\
 \text{newthyroid} & {$\mathbf{0.079}$} & 0.33 & {$\mathbf{0.12}$} & 0.14 & {$\mathbf{0.11}$} & 0.19 & {$\mathbf{0.084}$} & 0.15 & {$\mathbf{0.098}$} & 0.12 \\
 \text{pima} & {$\mathbf{0.29}$} & 0.38 & {$\mathbf{0.29}$} & 0.35 & {$\mathbf{0.27}$} & 0.28 & {$\mathbf{0.26}$} & 0.36 & {$\mathbf{0.27}$} & 0.29 \\
 \text{saheart} & {$\mathbf{0.37}$} & 0.39 & {$\mathbf{0.38}$} & 0.41 & {$\mathbf{0.29}$} & 0.40 & {$\mathbf{0.31}$} & 0.42 & {$\mathbf{0.29}$} & 0.36 \\
 \text{sonar} & {$\mathbf{0.20}$} & 0.46 & {$\mathbf{0.20}$} & 0.35 & 0.33 & {$\mathbf{0.32}$} & {$\mathbf{0.21}$} & 0.29 & 0.33 & {$\mathbf{0.28}$} \\
 \text{spectfheart} & {$\mathbf{0.23}$} & 1.0 & {$\mathbf{0.23}$} & 0.31 & {$\mathbf{0.20}$} & 0.21 & {$\mathbf{0.21}$} & 0.30 & {$\mathbf{0.20}$} & 0.25 \\
 \text{tae} & {$\mathbf{0.50}$} & 0.59 & {$\mathbf{0.49}$} & 0.58 & {$\mathbf{0.51}$} & 0.64 & {$\mathbf{0.53}$} & 0.54 & {$\mathbf{0.54}$} & 0.57 \\
 \text{vehicle} & {$\mathbf{0.30}$} & 0.49 & {$\mathbf{0.34}$} & 0.38 & {$\mathbf{0.38}$} & 0.45 & {$\mathbf{0.36}$} & 0.43 & {$\mathbf{0.38}$} & 0.45 \\
 \text{wdbc} & {$\mathbf{0.044}$} & 0.21 & {$\mathbf{0.060}$} & 0.077 & 0.11 & {$\mathbf{0.10}$} & {$\mathbf{0.072}$} & 0.11 & {$\mathbf{0.049}$} & 0.077 \\
 \text{wine} & {$\mathbf{0.056}$} & 0.29 & {$\mathbf{0.051}$} & 0.34 & {$\mathbf{0.062}$} & 0.29 & {$\mathbf{0.062}$} & 0.33 & {$\mathbf{0.062}$} & 0.34 \\
 \text{wisconsin} & {$\mathbf{0.032}$} & 1.0 & {$\mathbf{0.031}$} & 0.037 & 0.044 & {$\mathbf{0.031}$} & {$\mathbf{0.048}$} & 0.062 & {$\mathbf{0.029}$} & 0.029 \\
 \text{yeast} & {$\mathbf{0.48}$} & 0.55 & 0.48 & {$\mathbf{0.48}$} & 0.46 & {$\mathbf{0.42}$} & {$\mathbf{0.45}$} & 0.48 & 0.46 & {$\mathbf{0.42}$} \\
\hline
\end{tabular}}
\end{table*}

We tested the significance of the differences in the two experimental results, using Wilcoxon's signed rank tests. The Pareto-ranks of respective algorithms based {the error and selection rates from \tabref{tab:Error and Selection Rates Comparison} were compared with that of REPS with the alternative hypothesis that the average for REPS is smaller.}
For comparing the FSR errors, the values from the adjacent, corresponding columns in \tabref{tab:Comparison of FSR Error Rates} were taken and tested, respectively. 

The summary of the tests are shown in \tabref{tab:Comparison of Means and Signed Rank Test Results}. The first and the second rows show the $p$-values from the comparisons of the Pareto-ranks and the FSR errors, respectively. In the Pareto-ranks comparison test, the null hypotheses for all baselines but SSMA can be rejected with a high confidence. For the comparison of FSR errors, the null hypotheses for baselines other than SSMA and RMHC can be rejected with a high confidence. SSMA and RMHC were two best algorithms in \cite{Garcia:2012:PSN:2122272.2122582}, which indicates that REPS has a significant advantage against many prototype selection algorithms and is competitive with the state-of-the-art.

\begin{table*}[tb]
\caption{Summary of the Signed Rank Tests}
\label{tab:Comparison of Means and Signed Rank Test Results}
\centerline{
\begin{tabular}{clllll}
\hline
&\multicolumn{1}{c}{CCIS}&\multicolumn{1}{c}{FCNN}&\multicolumn{1}{c}{SSMA}&	\multicolumn{1}{c}{DROP3}&\multicolumn{1}{c}{RMHC}\\
\hline
 \text{PARETO} & 0.00019 & 4.6 $\times10^{-6}$ & 0.36 & 1.1$\times10^{-6}$ & 0.00043 \\
\text{FSR ERR}& 7.1$\times10^{-7}$ & 0.00011 & 0.39 & 9.6$\times10^{-6}$ & 0.11 \\
\hline
\end{tabular}
}

\end{table*}

\relocate{We compared between REPS and NoPS as other baseline methods do not take input of the dissimilarity matrix form. The error and the selection rates for the time series datasets are shown in \tabref{tab:Summary of Error and Selection Rates (Time Series)} in \apxref{apx:tables}. 
The problems where REPS improves the error rate are indicated by bold. 
Overall, REPS was able to reduce the number of prototypes while improving the error rates in three problems and maintaining the error rates in all others. The average selection rate over all problems was {0.88} with standard deviation {0.15}. The level of reduction was smaller than in the previous experiment due to smaller training set sizes. }

\relocate{We show the sensitivity analysis of the logarithm base $\beta$ of the soft maximum function. We compared the performances of REPS over the same collection of datasets while changing $\beta$ between 1 and 4. 
}
\relocate{\figref{fig:Sensitivity Analysis for beta} summarizes the result of the sensitivity analysis. The $y$-axis of the box-whisker chart indicates the value of the log odds ratio. It shows that the largest median is achieved with $\beta=2$, and the performance of REPS is robust with regards to the value of $\beta$.}

\relocate{
\begin{figure}[t]
\centerline{
\includegraphics[width=.35\textwidth]{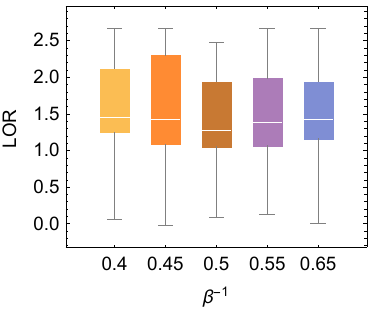}
}
\caption{Sensitivity Analysis for $\beta$}
\label{fig:Sensitivity Analysis for beta}
\end{figure}
}

\relocate{We present several case studies on the prototypes selected by the proposed method
in \apxref{apx:case_studies}.
These example cases demonstrated that the proposed method effectively addresses both the redundant patterns and noisy patterns without exploiting the explicit knowledge of the vector representation space.} 
 
\section{Conclusion}\label{sec:Conclusion}
This paper presented an extension of the nearest neighbor rule based on the adjustment of ranks to parametrize the prototype selection problem and also to approximate the violation of the rule over the training set. As a result, the problem is defined as discriminative learning in a sparse feature space. 
Our empirical results showed that it is competitive with the state-of-the-art prototype selection algorithm has the advantage over many other existing algorithms in multiple-measure comparisons.

\ifBibtex
\bibliography{/Users/ando/SkyDrive/Bibliography/All}
\bibliographystyle{siamplain}
\putbib
\else
\renewcommand{\url}[1]{}

\fi
\clearpage
\end{document}